# Mathematical Knowledge Graph-Driven Framework for Equation-Based Predictive and Reliable Additive Manufacturing

Yeongbin Cha

Department of Mechanical Engineering, Gachon University, Seongnam, Republic of Korea

dudqls7771@gachon.ac.kr

Namjung Kim (Corresponding Author)

Department of Mechanical Engineering, Gachon University, Seongnam, Republic of Korea

namjungk@gachon.ac.kr

https://orcid.org/0000-0002-2600-5921

## Abstract

Additive manufacturing (AM) relies critically on understanding and extrapolating process–property relationships; however, existing data-driven approaches remain limited by fragmented knowledge representations and unreliable extrapolation under sparse data conditions. In this study, we propose an ontology-guided, equation-centric framework that tightly integrates large language models (LLMs) with an additive manufacturing mathematical knowledge graph (AM-MKG) to enable reliable knowledge extraction and principled extrapolative modeling. By explicitly encoding equations, variables, assumptions, and their semantic relationships within a formal ontology, unstructured literature is transformed into machine-interpretable representations that support structured querying and reasoning. LLM-based equation generation is further conditioned on MKG-derived subgraphs, enforcing physically meaningful functional forms and mitigating non-physical or unstable extrapolation trends. To assess reliability beyond conventional predictive uncertainty, a confidence-aware extrapolation assessment is introduced, integrating extrapolation distance, statistical stability, and knowledge-graph-based physical consistency into a unified confidence score. Results demonstrate that ontology-guided extraction significantly improves the structural coherence and quantitative reliability of extracted knowledge, while subgraph-conditioned equation generation yields stable and physically consistent extrapolations compared to unguided LLM outputs. Overall, this work establishes a unified pipeline for ontology-driven knowledge representation, equation-centered reasoning, and confidence-based extrapolation assessment, highlighting the potential of knowledge-graph–augmented LLMs as reliable tools for extrapolative modeling in additive manufacturing.

# 1. Introduction

Rapid advancements in additive manufacturing (AM) have significantly broadened its applicability across diverse domains, including aerospace lightweight structures (Y. Huang et al., 2025; Zhang, Yan, et al., 2025), patient-specific biomedical implants (de la Rosa et al., 2024; Wu et al., 2025), energy devices (Khan et al., 2024), and architected mechanical metamaterials (Khalid et al., 2024). To sustain and further expand this applicability, several enabling technologies, such as efficient design strategies, material development, process monitoring, and computational modeling, have been continuously refined. In this context, understanding the influence of process parameters and systematically optimizing them remains one of the most critical challenges in additive manufacturing (Prabhakar et al., 2021). Traditionally, understanding and optimizing process-parameter relationships have relied on heuristic rules, expert experience, and extensive series of trial-and-error experiments or high-fidelity numerical simulations. While these approaches have yielded valuable insights, they are inherently constrained by high computational cost, long turnaround time, and limited scalability (Sani et al., 2024). Moreover, process knowledge is often optimized locally for specific machines, materials, and operating conditions. As a result, even when findings are reported in the literature, the accumulated knowledge remains fragmented and difficult to generalize or transfer across research groups and industrial settings.

Recent advances in artificial intelligence (AI), particularly large language models (LLMs), have attracted considerable attention due to their remarkable ability to extract, synthesize, and communicate knowledge from large-scale unstructured data (Naveed et al., 2025). These models have been explored for various manufacturing-related tasks, including process parameter recommendation (Ni et al., 2025), defect detection (K. Wang et al., 2025), too selection (Jia & Li, 2025), design-for-manufacturing guidance (Picard et al., 2025), and automated analysis of experimental logs and simulation reports (Dong et al., 2025). Such developments suggest a promising pathway toward data-driven and knowledge-assisted manufacturing workflows. Despite these successes, LLM face fundamental limitations that hinder their direct deployment in high-reliability engineering applications. A central concern is their tendency to generate hallucinations: outputs that appear plausible but are factually incorrect (L. Huang et al., 2025). This behavior arises because LLMs generate responses by predicting the most probable token sequences rather than explicitly retrieving or reasoning over verified physical knowledge. Consequently, the trustworthiness of LLM-based predictions becomes difficult to guarantee. In addition, LLM suffer from limited interpretability, as domain knowledge is encoded implicitly within high-dimensional model parameters, making it challenging to validate or audit the reasoning process (Singh et al., 2024). Because of this, their ways to obtain specific conclusions are inherently not understandable to human users.

To address the LLMs issues, one of the promising solutions is to utilize knowledge graphs (KGs) into LLM. KGs are a graph structure, consisting of nodes and edges, storing the relationships between nodes and structural information. It is known as effective and decisive manner of the way to store knowledge representation (Abu-Salih & Alotaibi, 2024). KGs are crucial for various applications as they offer accurate explicit knowledge without hallucinations. In addition, they are well known for their symbolic reasoning application, which helps the user to interpret their results (Liu et al., 2025). Moreover, the users can construct domain-specific KGs to provide precise and transferable domain-specific knowledge (Song et al., 2025). Despite these advantages, KGs have been limited to their static structures, and failing to handle incomplete data, making it difficult to apply real-world applications. To address these issues, the possibility of unifying LLM with KGs has attracted increasing attention from various areas (Ibrahim et al., 2024; Li et al., 2024). The KG-enhanced LLM, LLM-augmented KGs, and Synergized LLM and KG have extensively developed and utilized, expanding its applicability to various knowledge storing, interpretation, and engineering applications. More detailed information can be found in Ref. (Ibrahim et al., 2024; Liang et al., 2024; Yang et al., 2025).

In this study, we propose an ontology-guided, equation-centric framework that tightly integrates LLM with an additive manufacturing KGs to enable reliable knowledge extraction and extrapolative modeling. By explicitly structuring equations, variables, assumptions, and their semantic relationships within a formal ontology, the proposed approach transforms unstructured literature into machine-interpretable representations that can be systematically queried, conditioned, and extended. Building on this foundation, LLM-based equation generation is guided by mathematical KG-derived subgraphs, enforcing physically meaningful functional forms and mitigating non-physical or unstable extrapolation behavior commonly observed in unguided generation. Furthermore, a confidence-aware extrapolation assessment is introduced to distinguish predictive uncertainty from extrapolation reliability under limited data regimes. Together, these components establish a unified pipeline for ontology-driven knowledge representation, subgraph-conditioned equation reasoning, and principled extrapolation assessment.

## 2. Materials and Methods

### 2.1 Overall framework architecture

This study proposes an integrated framework that extracts equation-based knowledge from additive manufacturing (AM) literature, structures this knowledge into a hierarchical knowledge graph, and performs a reliable extrapolative prediction. The system is composed of six interconnected stages: (1) document preprocessing and chunking, which converts raw PDF documents into semantically segmented text units; (2) ontology-guided knowledge graph construction, which extracts entities and relations from text chunks according to a predefined schema; (3) hierarchical knowledge graph generation, which organizes extracted knowledge into multi-level abstraction layers; (4) multi-stage retrieval-augmented generation, which retrieves task-relevant subgraphs from the hierarchy; (5) equation-driven extrapolation using a LLM, which generates and fits candidate equations; and (6) confidence assessment, which evaluates prediction reliability based on statistical and physics-informed criteria. The detailed workflow is illustrated in Figure 1.

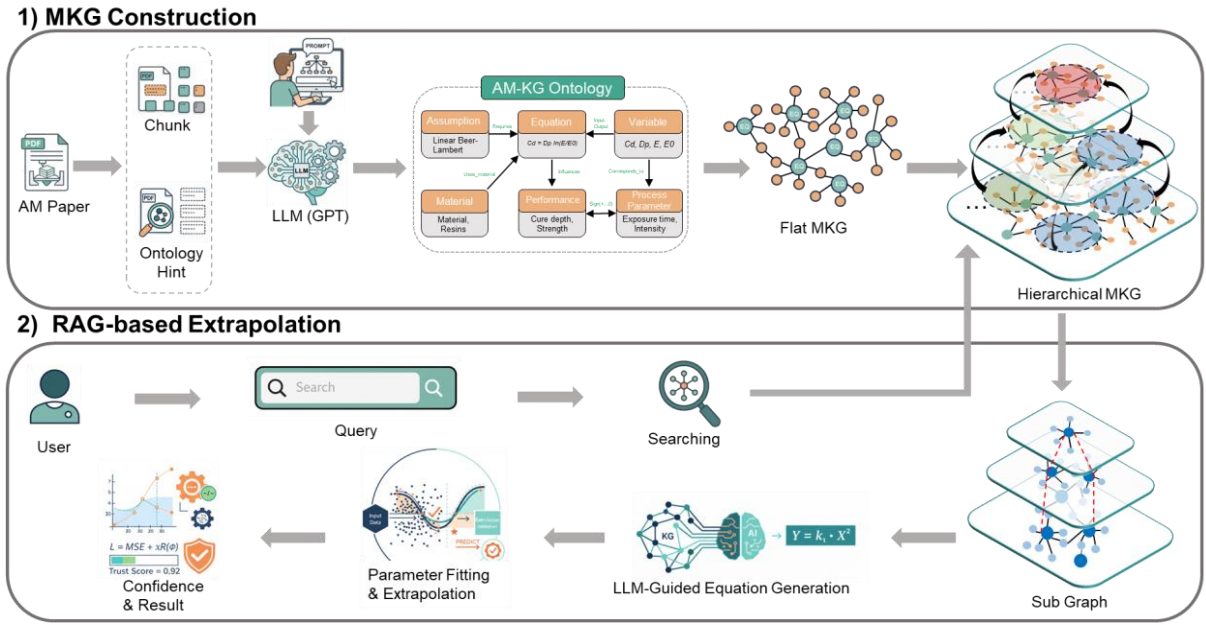

**Figure 1: Overall architecture of the proposed framework, consisting of (1) MKG construction and (2) RAG-based extrapolation.**

As shown in Figure 1, the proposed framework comprises two tightly coupled modules. The first module, the mathematical knowledge graph (MKG) construction, transforms unstructured AM research papers into a structured mathematical knowledge graph by extracting equations, variables, assumptions, and process–performance relationships, and organizing them into a hierarchical representation. This hierarchical MKG enables abstraction-aware access to equation-centered physical mechanisms while preserving the links to fine-grained entities.

The second module, RAG-based extrapolation, operates on the constructed hierarchical MKG to support physics-informed predictions beyond the observed data ranges. Given a user query, the system retrieves a task-relevant subgraph via a hierarchy-guided search. It then used an LLM to generate candidate governing equations using an LLM conditioned on the subgraph, and performs parameter fitting and extrapolation. A confidence assessment module then evaluates the prediction reliability by integrating the statistical evidence, prediction uncertainty, and physics-informed constraints derived from the knowledge graph.

By separating knowledge construction from reasoning and extrapolation, the proposed pipeline enables the systematic reuse of mathematical knowledge and supports domain knowledge-informed extrapolation beyond conventional data-driven prediction methods.

## 2.2 AM-MKG ontology design

**2.2.1 Ontology design rationale**

This study designed an ontology to consistently structure semantic relationships between mathematical equations and descriptive text within the AM domain. An ontology provides a formal knowledge representation that explicitly defines domain concepts and their interrelationships, enabling complex domain knowledge to be organized into its machine-interpretable form (Noy & McGuinness, 2001)Prior work on mathematical knowledge graphs has shown that explicitly representing mathematical components such as equations, variables, assumptions, and constraints strengthens the semantic linkage between symbolic expressions and textual descriptions, thereby improving interpretability and consistency during automated knowledge extraction.

This study follows a minimal task-driven ontology design strategy tailored specifically to equation-centered extrapolative modeling. Previous studies have reported that large and heterogeneous entity-type sets tend to increase extraction errors and hallucinations in LLM-based pipelines, whereas excessive prompt complexity can degrade extraction reliability (Maynez et al., 2020). Accordingly, we deliberately restrict the ontology to a few essential representational constructs required to support physics-informed equation interpretation and extrapolation.

**2.2.2 Competency-question–driven design**

To ensure that the ontology directly supports the reasoning operations required for downstream modeling, we formulated competency questions (CQs) using a task-driven design process, rather than a coverage-oriented approach. Following the methodology of Grüninger and Fox (1995), we first decompose the extrapolation task into its core analytical operations: (i) identifying the equation structure, (ii) interpreting parameter–performance relationships, (iii) determining applicability and validity regimes, and (iv) validating semantic consistency between equations and text. Subsequently, we derive CQs corresponding to each required inference. These CQs serve as testable criteria that specify the minimum knowledge that an ontology must represent.

Representative examples are provided to illustrate how competency questions guide the ontology design. For instance, the CQ "What are the input and output variables of a given equation?" requires equations to be represented as first-class entities and explicitly linked to variables through relations, such as has_input and has_output. Similarly, the CQ "Which process parameters influence a target performance metric, and in which direction?" necessitates the inclusion of causal influence relations between process parameters and performance entities along with directional sign annotations. These examples demonstrate the manner in which each CQ translates directly into specific representational requirements in an ontological schema.

### 2.2.3 Minimal ontology schema

Based on the derived CQs, the minimal AM-KG ontology comprises a compact set of core entity types, including equations, variables, assumptions, process parameters, performance, and phenomena, and a constrained set of semantic relations, such as has_input, has_output, influences, requires_assumption, and valid_in_regime. This minimal schema is sufficient to support equation-centered reasoning, semantic validation, and extrapolative modeling while avoiding unnecessary representational complexity that could hinder robust extraction. Unlike comprehensive domain ontologies aimed at exhaustive coverage, the competency questions defined were intentionally constrained to match only the representational requirements of equation-based extrapolation. Therefore, the CQ set functions as a requirements specification and a structural justification for the minimal ontology design adopted in this study. Appendix A provides a complete formal list of competency questions.

## 2.3 Knowledge graph construction pipeline

### 2.3.1 Preprocessing text and equation

Accurate extraction of mathematical expressions is essential because the behavior of the AM process is frequently described through equations that include symbolic notations, such as Greek letters, subscripts, and domain-specific variables. To preserve these expressions during document processing, we first applied a PDF-to-markdown conversion step designed to maintain the equation structure in the LaTeX form. This operation was implemented using *Nougat*, a vision transformer-based OCR model specialized for scientific document parsing that preserves mathematical notation in the LaTeX format (Blecher et al., 2023). When the conversion produced incomplete or fragmented equations, an additional text recovery stage was performed using a PyMuPDF-based extractor to ensure the complete retrieval of both textual and mathematical content (PyMuPDF, 2024).

Following equation acquisition, the text was segmented to preserve the semantic relationship between each equation and its surrounding explanatory context. Because separating an equation from the text that interprets it can distort the intended meaning of the governing relationship, we adopted an equation-centered chunking strategy rather than a fixed-length segmentation. Each equation and its adjacent context were combined into a single semantic unit, and token-based segmentation using tiktoken, OpenAI's byte-pair encoding tokenizer (OpenAI, 2023), was applied with a maximum length of 1024 tokens to avoid excessive chunk size while preventing semantic fragmentation (Ramakanth Bhat et al., 2025).

Ontology-guided preliminary hints were generated to improve the precision of the LLM extraction process. This step identifies candidate entities and relations that align with the ontology structure, such as process parameters, performance metrics, equation variables, and relational patterns, and presents them as contextual cues for LLM. The purpose of this preprocessing is not to produce final annotations, but to provide structured priors that reduce ambiguity and improve extraction reliability (Feng et al., 2024). Previous studies have shown that ontology-aligned prompting suppresses hallucinations and enhances extraction accuracy relative to free-text prompting (Sauka et al., 2025; Zeng et al., 2026). These preliminary results were validated and refined during the primary LLM-based extraction phase (Chepurova et al., 2024). Detailed procedures are provided in Appendix B.

### 2.3.2 LLM-based knowledge graph construction

Here, ontology-compliant knowledge graph triples are automatically constructed from each text chunk using structured LLM prompting. The extraction process leveraged the preprocessed text chunks and ontology-guided hints generated in the previous stage. This task was implemented using GPT-4o mini as the underlying language model (OpenAI 2024).

The extraction pipeline is explicitly divided into two sequential stages: entity and relation extraction. This separation reduces spurious entity generation and ensures that all extracted triples conform to the ontology-defined type constraints and semantic consistency requirements. By isolating entity identification from relation inference, the pipeline enforces tighter control over the generation behavior of the LLM and aligns with recent best practices in ontology-guided knowledge graph construction (Chepurova et al., 2024; Feng et al., 2024; Mo et al., 2025; Pan et al., 2025).

### 2.3.3 Entity extraction

During the entity extraction stage, ontology-guided hints were injected at the beginning of the prompt to provide structured prior knowledge before the model processed the input text. To avoid prompt overload and maintain formatting consistency, the number of hints was capped at fewer than 2,000 per ontology category (Y. Du et al., 2025).

The prompt explicitly instructs the model to verify the accuracy of the provided ontology hints, and adds missing entities only when supported by explicit textual evidence. This design encourages critical validation rather than blindly reproducing the initial candidate list. In recent ontology-guided entity extraction studies, candidate presentation and verification strategies have been shown to effectively reduce hallucinations and improve extraction accuracy(Feng et al., 2024; Pan et al., 2025).

The entity types are strictly defined using a JSONL output schema with explicit naming rules. Additionally, a few-shot learning strategy was employed by presenting two example pairs of sentences from real AM research papers and their corresponding annotated outputs. This enables the model to learn the expected extraction format and semantic scope unambiguously (Fei-Fei et al., 2006; Y. Wang et al., 2021). The detailed prompt configurations and examples are provided in Appendix C.

To ensure stability in the downstream relation extraction and consistency with the ontology, all extracted entities underwent rigorous normalization and validation. This process enforces the single source of truth (SSOT) principle, under which semantically equivalent concepts are consolidated into a unified representation (Papadakis et al., 2021). Entity names were normalized to a lower-case, underscore-separated snakecase format, with automatic conversion applied to uppercase letters and whitespace to eliminate redundant variants.

Because equation entities are crucial in representing physical knowledge, additional validation rules were applied. LaTeX expressions that lacked equality symbols or mathematical operators were excluded from being treated as valid equations (Rabbani et al., 2023). Entities sharing identical normalized names are merged to mitigate the duplication and incompleteness arising from chunk-level LLM extraction.

### 2.3.4 Relation extraction

In the relation-extraction stage, the semantic relations between entity pairs are inferred based on a sentence-level meaning structure. The LLM assigns relation types by leveraging the fixed list of candidate entities extracted in the previous stage, along with the ontology-defined constraints provided in the prompt. By fixing the entity set before relation generation, relations can only be produced between validated entities, a strategy shown to improve the extraction stability in LLM-driven knowledge graph construction (Feng et al., 2024).

To ensure schema compliance, the relation types were restricted to predicates defined in the ontology. The extracted relationships were then subjected to multiple validation checks before being finalized. For each triple, the subject and object are verified to exist within the normalized entity list. In addition, a structural constraint is enforced, whereby every equation entity must participate in a minimum of least one has_input and has_output relation, preventing isolated equation nodes within the graph.

This constraint-based validation strategy preemptively mitigates erroneous relation generation by LLM and ensures both the structural integrity and semantic coherence of the resulting knowledge graph.

### 2.3.5 Post-processing

Chunk-level LLM extraction can introduce global inconsistencies, including redundant entity creation under different surface forms and independently extracted fragmented relations across contexts. To address these issues and enforce ontology-wide consistency, corpus-level postprocessing was applied.

Duplicate entities are consolidated using Uniform Resource Identifiers (URIs), enabling semantically equivalent concepts extracted from different chunks to be merged into a single representative node. All available textual descriptions and evidence sources were aggregated to preserve provenance and semantic richness.

For equation entities, partial or variably structured LaTeX expressions can be extracted across chunk boundaries. During the merging, the most structurally complete and information-rich expression was selected as the canonical representation (Blecher et al., 2023; Lai et al., 2022). The relations are similarly merged across the chunks. When identical relations are extracted multiple times, their frequencies accumulate as weights, thereby providing a quantitative measure of evidential support. A weight of one indicates a single observation, whereas higher values indicate repeated confirmation across multiple contexts. To avoid redundancy, only one representative sentence was retained per relation, while the provenance information from all contributing chunks was preserved.

Finally, all the entities and relations were validated against ontology-defined constraints. Each equation entity must have at least one has_input and one has_output relation, and all influence relations must specify a directional sign (+, −, or 0). Additional assessments included endpoint consistency, regime–parameter mapping integrity, noncircularity of derived_from relations, and conflict detection among the assumptions. Entities or relations violating these constraints were excluded from the final graph.

This postprocessing stage reconciles distributed extractions across chunks, enforces ontology conformity, and ultimately enables the construction of a coherent and complete AM mathematical knowledge graph.

## 2.4 Hierarchical structuring and embedding

### 2.4.1 Hybrid entity embedding

Embedding is a representation-learning technique that maps discrete symbols, such as words, equations, and entities, into a low-dimensional continuous vector space, enabling semantic and structural similarities to be quantified through distance or angular measures(Bengio et al., 2003; Mikolov et al., 2013). In this study, ontology-derived entities and relations are embedded as continuous vectors to support semantic retrieval, retrieval-augmented generation (RAG), and hierarchical clustering (Lewis et al., 2020). To preserve the functional role of the equations in the downstream reasoning, an embedding process is designed such that the equation entities and their directly connected variables remain proximate in the embedding space.

To jointly encode for formulaic structure and natural language context, we designed a hybrid embedding module that transforms ontology-based entities and relations into vector representations by integrating both equation and textual information. Prior research on mathematical information retrieval has shown that modeling mathematical expressions and natural language separately while accounting for their distinct structural characteristics improves formula retrieval performance and cross-modal alignment between equations and text (Krstovski & Blei, 2018). Following this insight, our approach combines sentence transformer-based equation embedding with a text embedding model to construct a unified 1,920-dimensional vector representation for each entity and relation (Neelakantan et al., 2022; Reimers & Gurevych, 2019).

### 2.4.2 Embedding components: formula and text

Equation entities were first processed by translating their LaTeX representations into semantically interpretable English descriptions. This transformation is intended to expose core operators, symbols, and functional relationships in a form accessible to language-based embedding models. Mathematical symbols and operators are mapped to natural language equivalents, whereas extraneous formatting symbols are removed to reduce noise unrelated to formula semantics. The resulting descriptions were embedded using the all-MiniLM-L6-v2 sentence transformer model to produce a 384-dimensional vector (Reimers & Gurevych, 2019).

The vector was normalized to unit length to ensure stable scaling when combined with textual embeddings. For entities without LaTeX expressions, a zero vector is assigned to the formula component to maintain dimensional consistency across the entities. Although this approach does not aim to reconstruct a formal symbolic structure fully, it captures the dominant operators and variable relationships that are sufficient for equation-centered retrieval and clustering.

In parallel, the natural language attributes associated with each entity, including descriptions and metadata, are concatenated into a single text sequence and embedded using the OpenAI *text-embedding-3-small* model, which produces a 1,536-dimensional vector (OpenAI 2024). As with formula embeddings, text embeddings are normalized to unit length to ensure compatibility during hybrid vector construction.

### 2.4.3 Hybrid embedding

To integrate formulaic and textual representations into a single embedding space, formulas and text embeddings were combined using weighted vector fusion. In the additive manufacturing (AM) domain, most causal relations and semantic distinctions were conveyed through textual descriptions and ontology-defined categories, whereas equation entities constituted only a small fraction of the overall knowledge graph (approximately 4.8%, 48 of 994 nodes). Accordingly, textual embeddings are emphasized, whereas formula embeddings are incorporated into complementary structural signals.

Before fusion, both embedding vectors were L2-normalized to prevent higher-dimensional text embeddings from dominating the similarity calculations. In this study, text and formula embeddings are combined using a fixed 7:3 ratio (0.7 for text and 0.3 for formula). This weighting reflects the distribution of semantic information within the graph, where the majority of domain knowledge is encoded in text-based entities and mathematical formulas primarily serve as auxiliary structures that connect variables and physical relationships.

Using this structure-aware embedding, we observed that the cosine similarity between equation and output variable pairs was approximately 51% higher than that of randomly paired entities. This observation suggests that equation–variable relationships were effectively preserved in the embedding space. Therefore, formula-enriched RAG preferentially retrieves physically consistent equation–variable chains, thereby helping reduce hallucinations compared to text-only retrieval.

These hybrid embeddings serve as retrieval substrates and the basis for the hierarchical abstraction of the MKG. In the next section, we leverage this embedding space with ontology-defined relations to construct a multi-level hierarchical structure that enables mechanism-aware retrieval.

### 2.4.4 Hierarchical structure construction

Hierarchical clustering incrementally groups entities into semantically coherent clusters, thus forming a multi-level knowledge structure in which high-level abstractions coexist with low-level details. To mitigate the noise, redundancy, and exploratory inefficiency inherent in flat large-scale knowledge graphs, we adopted a hierarchical organization inspired by prior work, such as LeanRAG, which demonstrated improved retrieval efficiency and contextual coherence through hierarchical structuring (Zhang, Wu, et al., 2025).

A purely embedding-based clustering scheme risks overlooking the explicit structural relations encoded in the ontology, whereas structure-only clustering fails to group semantically related yet indirectly connected entities. To address this limitation, we proposed a hybrid hierarchical clustering framework that integrates ontology-defined relations with embedding-based semantic similarities (Krstovski & Blei, 2018; Misra et al., 2024). This framework combines equation-centric pre-clustering, Louvain-based community detection, and iterative re-embedding using LLM-generated summaries, while preserving connectivity with the original knowledge graph.

We construct a hybrid graph $G = (V, E_R \cup E_S)$, where $E_R$ denotes relation-based edges from ontology triples, and $E_S$ denotes similarity-based edges derived from hybrid embeddings. For a node pair $(i, j)$, the edge weight is defined as

$$\omega_{ij} = \begin{cases} n_{ij}\alpha, & (i,j) \in E_R, \\ (1-\alpha)\,sim(e_i, e_j), & (i,j) \in E_S \text{ and } (i,j) \notin E_R, \\ 0 & otherwise, \end{cases} \quad (1)$$

where $\alpha = 0.6$, and $n_{ij}$ is the number of observed relations between $i$ and $j$. Cosine similarity is computed as:

$$sim(e_i, e_j) = 1 - d_{cos}(e_i, e_j) \quad (2)$$

Similarity edges are added only among the $k = 10$ nearest neighbors whose similarity exceeds a threshold of $\theta_{sim} = 0.7$. This design prioritizes an explicit ontology structure when available, while using embedding-based similarity to complement sparsely connected regions. Appendix C provides the formal definitions.

All original entities and relations are preserved in Layer 0 (L0), ensuring that fine-grained knowledge remains accessible for downstream reasoning and drill-down retrieval. The higher layers retain explicit mapping to L0 to guarantee interpretability.

In Layer 1, ontology-type constraints are applied to form equation-centric clusters representing the local physical modeling units. For each equation, the adjacent variables and assumptions are identified as follows:

$$N(q) = \{v \mid (q,v) \in E_R,\ v \in V_{var} \cup V_{assump}\} \tag{3}$$

An equation-centered cluster is then defined as:

$$C_q = \{q\} \cup N(q), \tag{4}$$

and created whenever $|C_q| \geq 3$. This constraint ensures that the equations remain grouped with their modeling components rather than dispersed across heterogeneous communities.

Entities not assigned to equation-centric clusters were grouped by applying the Louvain algorithm to the hybrid graph (Blondel et al., 2008). For higher layers (layers 2 and above), clusters from the previous layer were treated as units for another round of Louvain clustering. For each cluster $C_k$, we computed the centroid embedding as

$$c_k = \frac{1}{|C_k|} \sum_{i \in C_k} e_i, \tag{5}$$

and generated a one-sentence semantic summary $s_k$ using GPT-4o mini(OpenAI, 2024). The new embedding for the next layer is constructed as

$$e_k^{(l+1)} = \begin{bmatrix} 0_{384} \\ TextEmbed(s_k) \end{bmatrix}, \tag{6}$$

yielding a 1,920-dimensional hybrid representation.

Inter-cluster relations were aggregated by identifying the dominant relation type between cluster pairs and recording their frequency as an edge weight. Additional LLM-generated summaries are attached to critical relations, such as influences and has_input.

This iterative process continues until the number of clusters falls below a stopping threshold $\theta_{stop} = 20$. Applying this procedure to our 994-entity knowledge graph resulted in 211 equation-aware clusters in Layer 1, which were further abstracted into 25 higher-level clusters. The constructed hierarchy preserves both the ontology structure and embedding-based semantic similarity, distinguishing it from purely data-driven clustering approaches.

As part of the hierarchical construction, the hyperparameters $\theta_{sim}$ and $\alpha$ were determined through sensitivity analysis focusing on hierarchical convergence, cluster granularity, and computational efficiency. The similarity threshold $\theta_{sim}$ directly controls graph density by regulating the inclusion of similarity-based edges, and experiments were conducted over $\theta_{sim} \in \{0.5, 0.6, 0.7, 0.8, 0.9\}$. The results showed that lower values led to overly coarse clusters owing to the excessive merging of semantically distinct entities, whereas higher values caused over-fragmentation and degraded convergence at higher hierarchy levels.

In contrast, $\theta_{sim} = 0.7$ produced 325 Layer-1 clusters that reliably converged to 20 Layer-2 clusters, satisfying the stopping criterion ($\theta_{stop} = 20$) with a compression ratio of 6.2%. The relation weight $\alpha$ primarily influenced the degree of abstraction at higher layers rather than the base clustering structure. Across $\alpha \in \{0.4, 0.6, 0.8\}$, the number of Layer-1 clusters remained nearly constant (CV < 0.2%), indicating structural robustness. Among these, $\alpha = 0.6$ achieved the fastest runtime while consistently yielding a stable two-layer hierarchy. Accordingly, $\theta_{sim} = 0.7$ and $\alpha = 0.6$ were adopted as the final configuration.

## 2.5 MKG-RAG for equation retrieval

### 2.5.1 Overview and query conditioning

To retrieve physically meaningful modeling knowledge from a constructed MKG, we employed a RAG framework that operates directly on structured mathematical entities rather than on unstructured text (Lewis et al., 2020). This module aims to extract a compact subgraph by selectively retrieving and expanding ontology-constrained entities that capture coherent physical mechanisms containing equations, variables, assumptions, and related physical contexts relevant to a given modeling task. The retrieval process was conditioned on a set of user-specified input variables.

$$X = \{x_1, \cdots, x_m\} \tag{7}$$

and optional target variable $y$. When a natural language query was provided, it was used directly. Otherwise, the system generates a deterministic, domain-grounded query using the following template: "equations and mechanisms $x_1, \cdots, x_m, y$ in additive manufacturing". This formulation explicitly embeds the symbolic variable identifiers within the domain context, ensuring consistent retrieval behavior across different modeling scenarios.

### 2.5.2 Hybrid retrieval with hierarchical mechanism expansion

Given the constructed query, a semantic search is performed in a hybrid embedding space that jointly encodes entity descriptions, mathematical expressions, and relational contexts, as described in Section 2.5. This hybrid representation promotes the retrieval of physically consistent equation–variable, associations, rather than isolated textually similar entities. The top-$k$ nearest entities ($k = 10$ in all experiments) were retrieved based on cosine similarity. To prevent critical variables from being omitted owing to embedding noise or notation mismatch, all specified input and target variables were explicitly included in the candidate set.

To mitigate the limitations of flat-embedding-based retrieval, MKG-RAG employs a hierarchy-guided expansion strategy inspired by LeanRAG (Zhang, Wu, et al., 2025). The system identifies the lowest common ancestor (LCA) cluster and deepest cluster in the hierarchy containing all retrieved entities using the hierarchical clustering constructed in Section 2.6. Because higher-level clusters correspond to abstracted physical mechanisms encoded during hierarchical construction, the candidate subgraph is expanded to include all entities within the LCA cluster and its descendant clusters, along with the direct graph neighbors of the retrieved entities and all ontology-defined relations among them. This expansion balances the retrieval completeness and physical coherence without uncontrolled graph growth.

Importantly, the proposed retrieval strategy does not assume explicit governing equations. When no such equation exists in the retrieved subgraph, the MKG-RAG framework returns a mechanism-level representation composed of variables, physical phenomena, and ontology-defined relations, thereby providing a structured basis for subsequent equation composition.

### 2.5.3 Ontology-aware equation role assignment

The retrieved equations did not contribute equally to the modeling task. To distinguish the functional roles, each equation in the extracted subgraph was automatically annotated using ontology-defined relations(Da Cruz et al., 2025). Equations directly linked to the target variable through relations such as *has_output* are labeled as target equations. Equations that depend explicitly on one or more input variables through relations, such as *has_input* or *influences,* are labeled as input-related equations. Additional equations that are indirectly connected through equation–to-equation dependencies are labeled as supporting equations. Role assignment was performed by first scanning direct equation–variable relations, followed by a constrained graph traversal over equation adjacency to include mechanically relevant supporting equations without uncontrolled subgraph expansion. These role annotations were later used to prioritize physically meaningful equation chains during LLM-based equation composition.

### 2.5.4 Structured subgraph output

The final output of the MKG-RAG module is a structured JSON-serialized subgraph that includes equations, variables, assumptions, regimes, materials, and physical phenomena along with ontology-constrained relations and hierarchy-level summaries (Zhang, Wu, et al., 2025). Each equation is annotated with its functional role and explicit variable dependencies, and the subgraph is accompanied by metadata describing the entity distributions and role assignments. By representing equations as explicit mathematical entities and preserving physical assumptions and validity regimes as first-class components, the resulting subgraph provides a structured and interpretable input, in which equation candidates, variable dependencies, and physical assumptions are explicitly exposed for downstream equation composition and extrapolative modeling.

## 2.6 Equation discovery and extrapolation modeling

### 2.6.1 Overview

Given a task-specific subgraph retrieved by the MKG-RAG module, we perform equation composition and extrapolative modeling using a pipeline that integrated LLM–based candidate generation with symbolic processing and parameter estimation. This stage aims to construct closed-form, physically interpretable equations that relate specified input variables to a target variable, and to calibrate these equations using observed data for extrapolative analysis.

Rather than treating equation generation as a purely text-driven task, the proposed pipeline conditions LLM on a structured subgraph containing variables, equations, physical assumptions, validity regimes, and hierarchical summaries (M. Du et al., 2024; Lewis et al., 2020). This design ensures that candidate equations are generated within a constrained physical and semantic context, thereby enabling subsequent symbolic parsing and numerical fitting.

### 2.6.2 Ontology-constrained equation generation with LLM

Candidate equations were generated using an LLM-based equation conditioned on the retrieved MKG-RAG subgraphs. The composer adopts a two-stage prompting strategy that separates system-level generation constraints from task-specific contextual information, while enforcing JSON-only outputs to ensure downstream machine interpretability.

The system promptly defines the mandatory constraints that every generated equation must satisfy (OpenAI et al., 2023). First, the candidate equations must explicitly express the target variable as a function of the specified input variables. Secondly, using variables, symbols, or physical concepts that are not present in the retrieved MKG subgraph is prohibited, which serves as an anti-hallucination constraint (Ji et al., 2023). Third, all free parameters must follow a fixed naming convention (e.g., $k_1, k_2$) to enable reliable symbolic parsing and automatic parameter fitting in the subsequent stages. Finally, the output must conform to a structured JSON schema containing LaTeX-formatted equations restricted to standard mathematical operators, such as fractions, natural logarithms, and exponential functions.

The user-level prompt provided a serialized, human-readable representation of the retrieved MKG-RAG subgraph, including equation entities, variable descriptions, ontology relations, physical assumptions, validity regimes, and hierarchy-level summaries. The task specification explicitly instructs the LLM to generate $M$ candidate equations that express the target variable as a function of specified input variables, with each candidate returned as a single-line LaTeX equation. This structured prompting strategy yielded a finite set of ontology-grounded symbolic equation candidates suitable for downstream symbolic and numerical processing (Cranmer et al., 2020). The complete algorithmic procedure for the ontology-constrained equation generation is summarized in Algorithm E.1 in Appendix E.

### 2.6.3 Symbolic parsing and construction of fitting functions

Each generated candidate equation is processed using a symbolic parser that converts its LaTeX representation into a symbolic expression. During parsing, mathematical operators and functions are normalized to ensure compatibility with symbolic evaluation. For example, logarithmic expressions are consistently mapped to natural logarithms to avoid ambiguity arising from base specifications.

From the parsed symbolic expression, the right side is decomposed into independent and free variables. A callable function compatible with non-linear least squares fitting was then constructed in the following form:

$$y = f(x_1, \cdots, x_m; \theta) \tag{8}$$

where $\theta$ denotes the set of unknown parameters. This function wrapper supports both univariate and multivariate inputs and allows the direct evaluation of vectorized data. Candidate equations that failed symbolic parsing or contained unresolved symbols were discarded at this stage, ensuring that only mathematically valid expressions proceeded to numerical fitting.

### 2.6.4 Parameter estimation and uncertainty quantification

For each valid candidate equation, the parameter values were estimated using nonlinear least-squares optimization. The initial parameter values were set to unity, unless otherwise specified, and an increased iteration budget was used to reduce premature convergence failures. The model performance was evaluated using the coefficient of determination ($R^2$) and root mean squared error (RMSE) computed from the residuals between the observed and predicted values (Hastie et al., 2009). Parameter uncertainty was quantified using the covariance matrix returned by the optimizer, from which standard errors and 95% confidence intervals were derived based on the student's t-distribution with appropriate degrees of freedom. This procedure yielded both calibrated parameter values and uncertainty estimates for subsequent comparative evaluations.

## 2.7 Confidence scoring framework

To quantitatively assess the reliability of the model predictions beyond the experimental data range, this study proposes an extrapolation confidence scoring framework (Storm et al., 2017). The framework integrates the extrapolation distance, statistical model quality, prediction uncertainty, and physics-informed constraints derived from an AM knowledge graph (AM-MKG) (R. Wang & Cheung, 2023).

The proposed method assigns a continuous scalar confidence score to each extrapolated prediction. This score collectively reflects the strength of empirical support, the distance from the training domain, and the consistency with domain knowledge encoded in the graph. Confidence analysis is performed on experimentally acquired data, extrapolation evaluation points, and analytically fitted models that are calibrated using nonlinear least-squares optimization (Vugrin et al., 2007). The final confidence score integrates extrapolation distance, statistical evidence, and physics-based consistency into a quantitative indicator.

### 2.7.1 Extrapolation distance and distance-based confidence

Extrapolation distance quantifies the extent to which an evaluation point lies outside the training domain (Chen, 2015). For variable $x$ with training $[x_{min}, x_{max}]$, the extrapolation distance is defined as follows:

$$d(x) = \begin{cases} x - x_{max}, & x > x_{max} \\ x_{min} - x, & x < x_{min} \\ 0 & otherwise. \end{cases} \tag{9}$$

For multivariate extrapolation, the extrapolation distances computed for the individual variables were normalized by their respective training ranges and combined using a Euclidean formulation to obtain a single normalized extrapolation distance.

$$d_{norm} = \sqrt{\sum_i \left( \frac{d(x_i)}{x_{max,i} - x_{min,i}} \right)^2} \tag{10}$$

The normalized extrapolation distance was then mapped to a confidence score using an exponential decay function.

$$S_{dist} = \exp(-\alpha_d \cdot d_{norm}) \tag{11}$$

where $\alpha_d$ is a parameter that controls the rate at which confidence decreases as the extrapolation distance increases. This formulation ensures that predictions within the training domain incur no penalty, whereas confidence decreases smoothly and monotonically in the extrapolation regions.

### 2.7.2 Confidence based on model fit and prediction uncertainty

Statistical confidence reflects both the quality of model calibration and the stability of the predictions. The model fit quality was quantified using $R^2$ computed on the training data and was constrained to the range [0,1],

$$S_{fit} = clip(R^2, 0, 1) \tag{12}$$

Prediction uncertainty was estimated using a bootstrap resampling procedure (DiCiccio & Efron, 1996). The training data were repeatedly resampled with replacements, and the model was refitted for each resampled dataset to generate a predictive distribution. From this distribution, a 95% prediction confidence interval was computed, and the uncertainty score was defined as

$$S_{uncertainty} = 1 - \min\left(1, \frac{CI\ width}{\sigma_y}\right) \tag{13}$$

where $\sigma_y$ denotes the standard deviation of the target variable in the training dataset. This normalization expresses the prediction uncertainty relative to the intrinsic scale of the response variable.

The statistical confidence component was computed as the arithmetic mean of the fit quality score and uncertainty score:

$$S_{start} = \frac{S_{fit} + S_{uncertainty}}{2} \tag{14}$$

This formulation balances model accuracy and predictive stability without introducing additional assumptions.

### 2.7.3 Physics- and knowledge-graph-based confidence

In addition to the geometric distance and statistical measures, the proposed framework incorporates physics-informed constraints derived from the additive manufacturing knowledge graphs. When a process-specific subgraph is available, the modeling assumptions associated with the governing equations are extracted, and a confidence penalty is applied based on the number of assumptions involved.

$$S_{assumption} = 1 - \min(0.3, 0.05 \times N_{assumptions}). \tag{15}$$

Furthermore, the AM knowledge graph encodes the causal influence relations between the process parameters and performance metrics, including the expected sign of influence. In this study, the signs of the fitted model parameters were compared with the influence signs specified in the knowledge graph to assess consistency.

$$S_{sign} = 1 - \frac{N_{mismatch}}{N_{matched}}. \tag{16}$$

The physics- and knowledge-graph-based confidence component is computed multiplicatively as

$$S_{phys} = S_{regime} \times S_{assumption} \times S_{sign}, \tag{17}$$

where $S_{regime} = 1.0$ in the current implementation, as the explicit process regime validation has not yet been incorporated.

### 2.7.4 Final confidence score and interpretation

The final extrapolation confidence score was computed as a weighted linear combination of all confidence components (Storm et al., 2017).

$$C_{total} = \alpha S_{dist} + \beta S_{stat} + \gamma S_{phys} + \delta S_{uncertainty}. \tag{18}$$

where $\alpha, \beta, \gamma,$ and $\delta$ are user-defined weights that control the relative contribution of each component. The resulting confidence score lies in the range [0, 1], and represents a relative measure of extrapolation reliability. It is not intended as a strict criterion for extrapolation validity but rather as a quantitative indicator for comparing extrapolated predictions generated under different models or operating conditions.

A larger weight was assigned to the distance-based component to reflect the widely accepted principle that the prediction reliability degrades as inputs move farther outside the training domain, making the extrapolation distance a primary risk factor for reliability assessment (Weaver & Gleeson, 2008).

In this study, the weights were set to α = 0.4, β = 0.3, γ = 0.2, and δ = 0.1, reflecting the relative importance of extrapolation distance, statistical reliability, physics-based consistency, and prediction uncertainty, respectively.

## 2.8 Data source and experimental setup

### 2.8.1 Source papers for AM-MKG construction

The source papers used to construct the additive manufacturing mathematical knowledge graph (AM-MKG) were selected to support equation-centered knowledge extraction rather than comprehensive literature coverage. Twenty papers were included based on the following criteria: The goal of this selection was to obtain mathematically explicit and quantitatively interpretable descriptions of the relationships between the process parameters, physical phenomena, and performance outcomes relevant to vat photopolymerization.

Accordingly, the selected papers provided one or more of the following types of information: (i) explicit mathematical formulations with clearly defined input and output variables; (ii) quantitative relationships linking controllable process parameters to measurable performance metrics; and (iii) physical assumptions or validity conditions underlying photopolymerization models. These studies collectively supply the minimum but sufficient knowledge required to populate AM-MKG entities, such as equations, variables, assumptions, and parameter–performance relations.

This role-oriented selection strategy ensures that the resulting AM-MKG captures the essential structure required for equation-centered reasoning and extrapolation, without introducing unnecessary redundancy or representational complexity.

A full list of the 20 source papers, including bibliographic information and DOIs, is provided in Appendix D to ensure the reproducibility of the AM-MKG construction corpus.

We emphasize that this corpus is role-curated to cover equation-centric knowledge needs rather than intended as an exhaustive systematic review.

### 2.8.2 Experimental dataset for extrapolation validation

For extrapolation validation, we used the experimental working-curve data reported in an interlaboratory study on vat photopolymerization by (Kolibaba et al., 2024). This study

- 23 -

measured the cure depth as a function of radiant exposure under controlled optical conditions, and multiple datasets were reported to reflect the variability across experimental procedures. From this collection, we selected a single dataset (Dataset22) for the controlled case study. Dataset_22 reports the cure depth measurements over a broad and continuous range of radiant exposure values while maintaining fixed wavelength and intensity conditions. Because radiant exposure is the governing independent variable in Jacobs-type working curve formulations, this dataset allows a clear separation between the fitting range and the out-of-range exposure condition used for extrapolation assessment.

Each radiant exposure level included replicate measurements of cure depth, which were retained to reflect experimental variability and evaluate the robustness of the equation fitting and extrapolation confidence estimation. The extrapolation was evaluated at a single exposure level beyond the fitting range, serving as a controlled stress test rather than a comprehensive benchmark.

This experimental design enables a focused analysis of the equation-based extrapolation behavior within the proposed AM-MKG-driven framework, whereas broader generalization across datasets is left for future investigation.

## 3. Results

### 3.1 Analysis of constructed knowledge graph

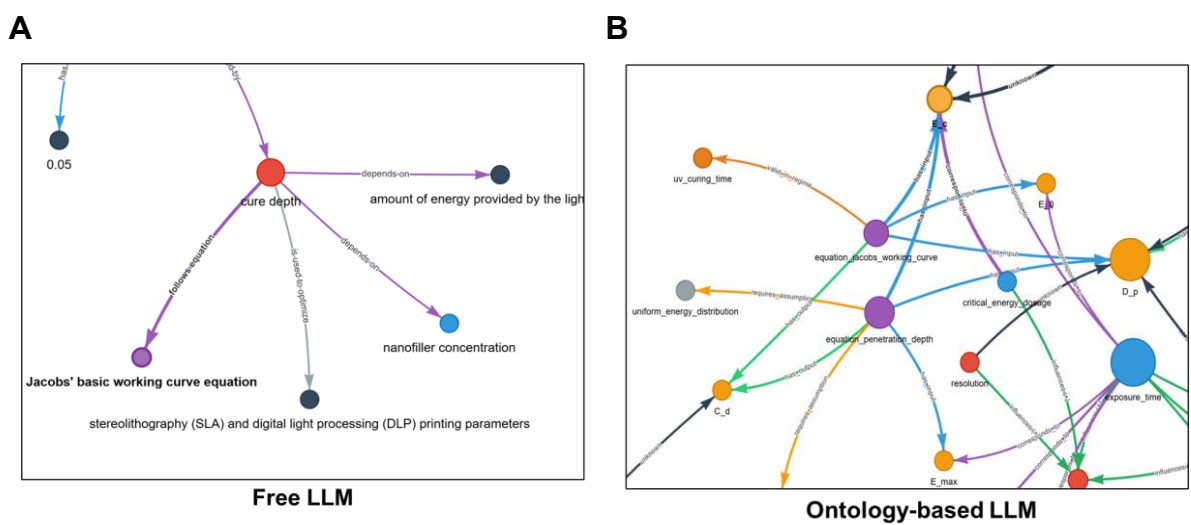

**Figure 2. Knowledge graphs generated by free LLM extraction and ontology-guided extraction. (A) Without ontology constraints. (B) With AM-MKG ontology.**

To analyze the impact of ontology constraints on equation-centric knowledge extraction, Figure 2 compares the knowledge graphs generated by free LLM-based extraction and ontology-guided extraction. As shown in Figure 2A, the knowledge graph produced by the unconstrained LLM extraction exhibits connectivity patterns that largely reflect the lexical co-occurrence among the terms in the source text. Process parameters, equation variables, and performance-related concepts were connected without clear semantic role differentiation. The variables in mathematical expressions are not explicitly structured as inputs or outputs of the equations, and the physical assumptions required for the validity of the equations are not explicitly represented. By contrast, the ontology-guided extraction result shown in Figure 2B demonstrates a more structured representation enabled by the predefined AM-MKG ontology. The equations, variables, process parameters, and physical assumptions are systematically distinguished and consistently connected. In particular, the Jacobs working curve is represented as an independent equation entity with has_input and has_output relations that explicitly that specify the dependencies between the input and output variables. Auxiliary variables and process-related conditions were connected using a consistent semantic framework.

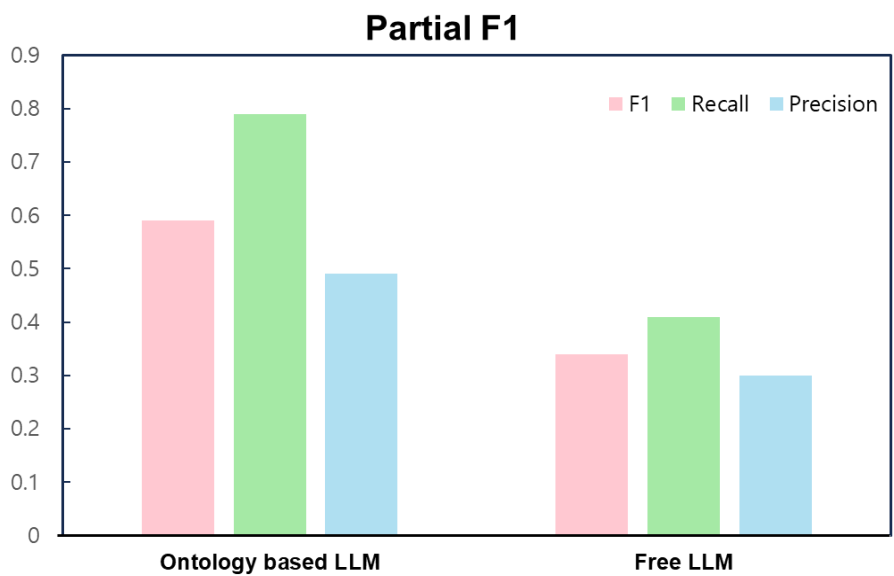

**Figure 3. Quantitative comparison of ontology-guided and free LLM-based extraction using the Partial F1 metric**

Figure 3 shows a quantitative comparison of the extraction performances. The Partial F1 metric was employed to account for common expression variations in the LLM-based extraction. Such variations include alternative surface realizations with equivalent meaning (e.g., exposure_time vs. UV exposure time), minor structural differences in relation expressions, and equivalent reformulations of variable names, which would be overly penalized under strict exact-match evaluation. The extraction performance was evaluated by comparison with a manually curated reference knowledge graph, and the precision, recall, and partial F1 were reported following the WebNLG+ evaluation protocol (Ferreira et al., 2020).

As shown in Figure 3, ontology-guided LLM extraction consistently outperformed free LLM extraction across all reported metrics. An improvement in precision indicates a reduction in spurious relations, whereas an increase in recall reflects an enhanced coverage of semantically valid relations. Consequently, the overall Partial F1 score was significantly higher for the ontology-guided extraction.

Altogether, these results indicate that ontology constraints improve the structural consistency and quantitative robustness of the extraction.

## 3.2 Extrapolation behavior with/without MKG guidance

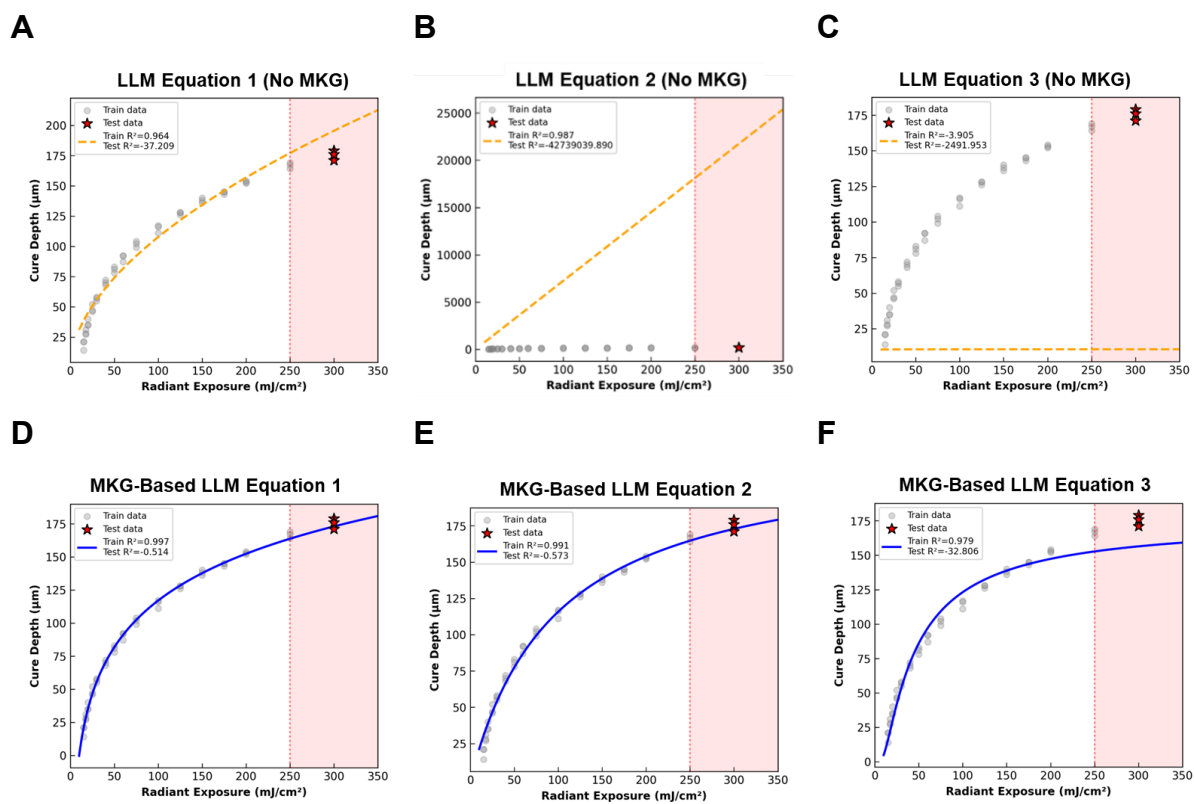

**Figure 4. Extrapolative behavior of equations generated with and without MKG-based subgraph guidance. (A–C) Without guidance. (D–F) With guidance. Shaded region: extrapolation regime.**

Figure 4 shows comparisons of the equations generated by LLM with and without conditioning on the MKG-based subgraphs. Without subgraph guidance (Figure 4A–C), the generated equations fit the training data but exhibit unstable behavior in the extrapolation regime, including divergence or numerical instability beyond the training range. In contrast, when MKG-based subgraphs were provided as contextual guidance (Figure 4D–F), the generated equations followed saturating functional forms, which is consistent with the Jacobs working curve behavior commonly observed in vat photopolymerization. These equations maintained stable and physically plausible trends in the extrapolation region.

This comparison suggests that the MKG-based subgraph guidance constrains equation generation in a manner that mitigates unstable extrapolative behavior.

## 3.3 Equation generation and predictive performance

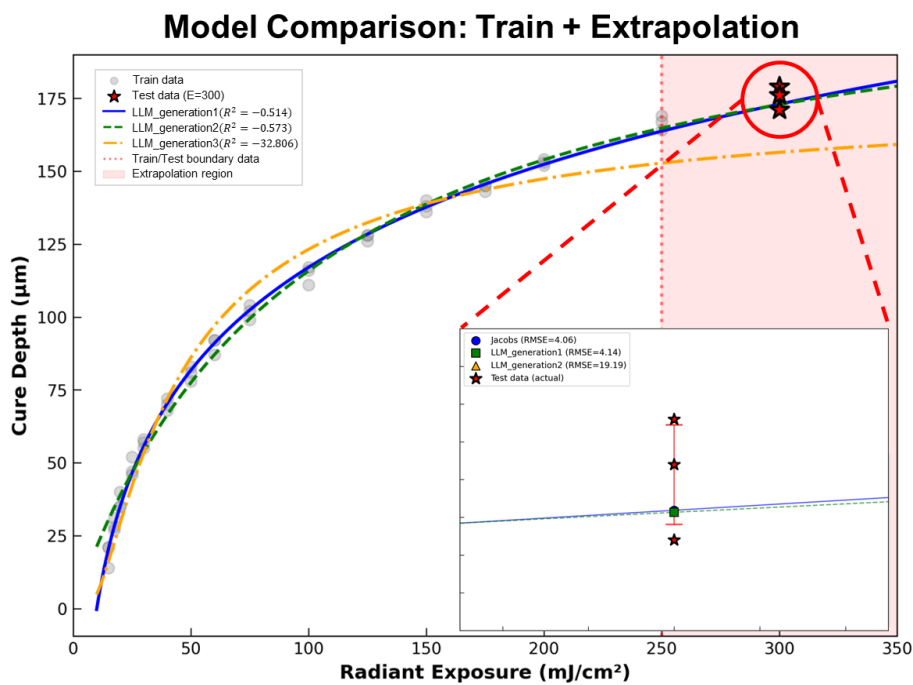

**Figure 5. Subgraph-guided equations for cure depth prediction. Shaded region: extrapolation regime.**

For clarity, the equations generated by LLM under MKG-based subgraph guidance in Figure 5 are summarized as follows:

(1) Jacobian working curve: $C_d = K_1 \ln\left(\frac{E}{K_2}\right)$;

(2) Ration form, $C_d = K_1 \frac{E}{(K_2+E)}$;

(3) Exponential form, $C_d = K_1 \exp\left(-\frac{K_2}{E}\right)$,

where $C_d$ denotes the cure depth and $E$ represents the radiant exposure.

Figure 5 presents the equations generated by LLM for modeling the resin cure depth as a function of radiant exposure using MKG-based subgraph guidance. The query subgraph explicitly includes the Jacobian working curve as a canonical physical model for vat photopolymerization. Therefore, the LLM reproduces the Jacobs equation and generates alternative equations that exhibit equivalent physical behavior.

As shown in Figure 5, all generated equations achieve comparable fitting performance within the training range (radiant exposure ≤ 250 mJ/cm²). Despite the differences in functional form (logarithmic, rational, and exponential), the equations consistently preserved the monotonic saturation behavior with increasing exposure. This indicates that the generated equations share a common underlying physical structure rather than representing arbitrary curve fitting.

The extrapolation results at 300 mJ/cm² further illustrated this consistency. Although the equations differ mathematically, their predictions remain within a physically plausible range and follow trends consistent with those of established photopolymerization models. These results show that MKG-based subgraph guidance enables the generation of multiple structurally distinct yet physically consistent equations, supporting stable and physically plausible extrapolations beyond the training domain.

### 3.4 Prediction uncertainty in the extrapolation regime

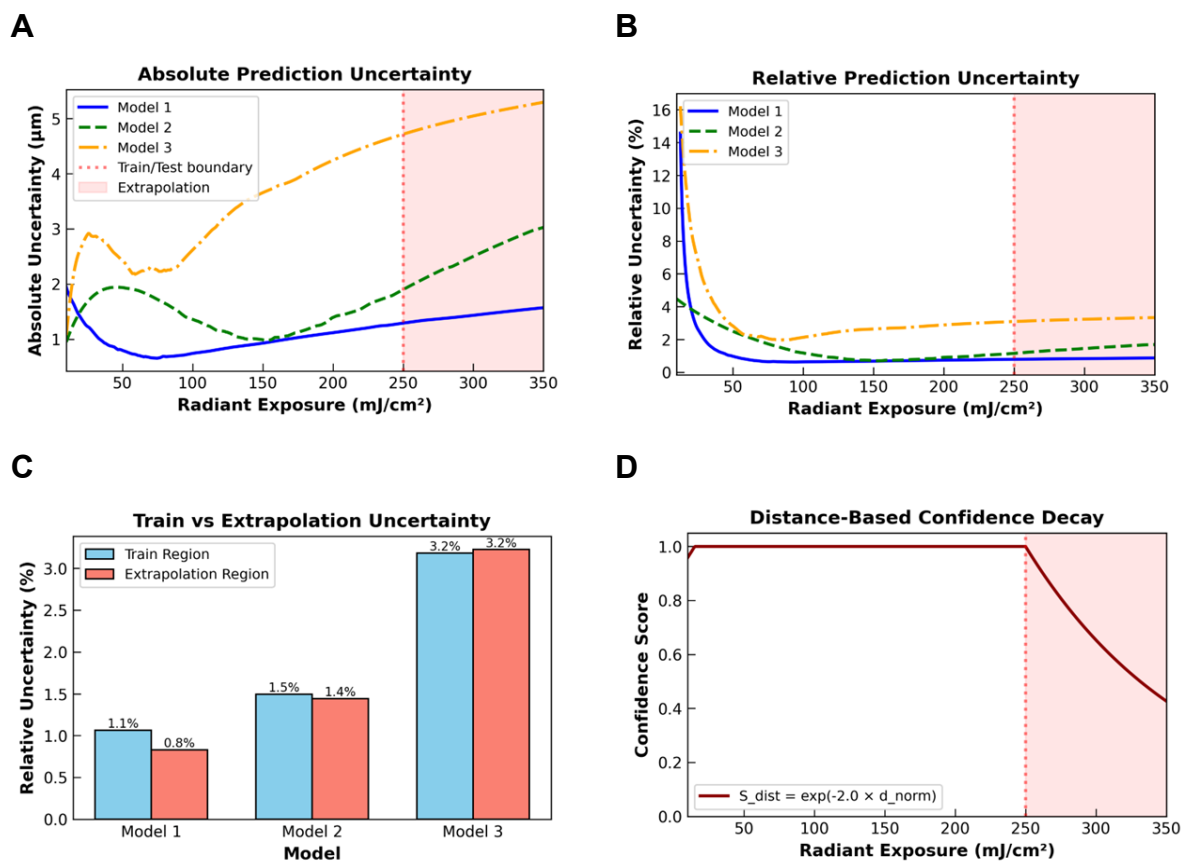

**Figure 6. Prediction uncertainty across training and extrapolation regimes. (A–B) Absolute and relative uncertainty. (C) Average relative uncertainty. (D) Distance-based confidence decay.**

Figure 6 shows comparisons of the prediction uncertainty behaviors of the equation-based models within the training and extrapolation regimes. Figures 6A and 6B present the absolute and relative prediction uncertainties, respectively, as functions of radiant exposure. The red dashed line indicates the boundary between the training and extrapolation domains, and the shaded region denotes the extrapolation regime.

Within the training domain, all models exhibited relatively low and stable uncertainty levels. However, as the radiant exposure extended beyond the training range, clear differences in uncertainty behavior began to emerge across the models. Model 1 shows a gradual and controlled increase in both absolute and relative uncertainties in the extrapolation regime, whereas Models 2 and 3 exhibit a more pronounced increase in uncertainty with increasing extrapolation distance. In particular, Model 3 demonstrates a non-linear amplification of the uncertainty beyond the training boundary, indicating structural instability when applied outside the calibrated data range.

Figure 6C summarizes the average relative prediction uncertainties in the training and extrapolation regimes. Although the uncertainty levels were comparable across models within the training domain, they became distinctly separated in the extrapolation regime. This separation highlights that extrapolative stability is governed not only by the goodness-of-fit within the training data but also by the structural properties of the underlying equations.

These quantitative results reinforce the qualitative extrapolation behaviors observed in Sections 4.2 and 4.3, demonstrating that the equations generated under the MKG-based subgraph guidance exhibit comparatively stable uncertainty characteristics in the extrapolation regime.

## 3.4 Confidence assessment of extrapolations

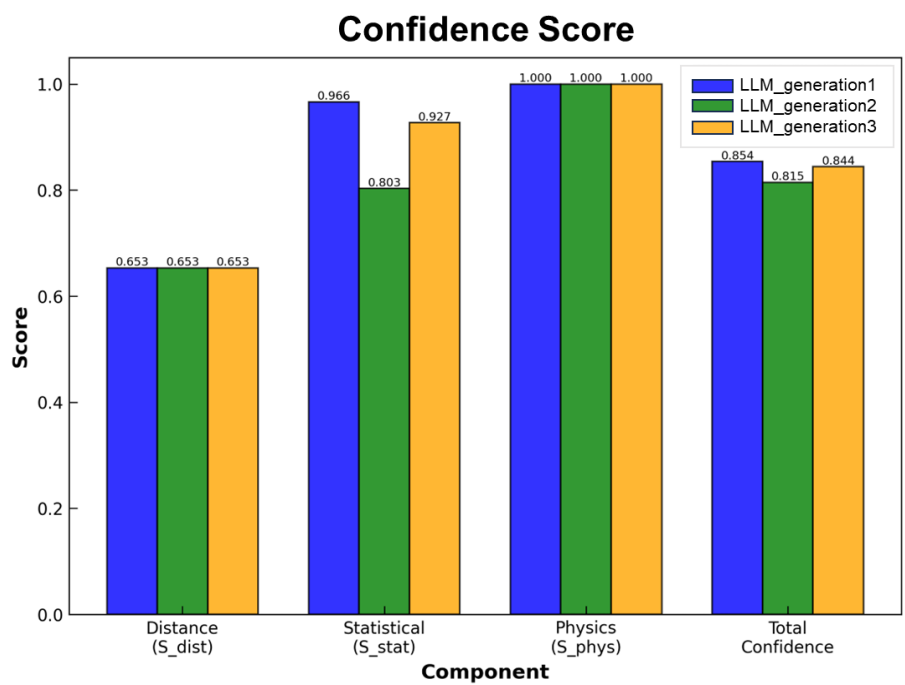

**Figure 7. Breakdown of confidence score components for extrapolative assessment.**

Figure 6D and 7 illustrate the behavior and relative contributions of the individual confidence components in the extrapolation regime. Figure 6D shows the distance-based confidence decay component as a function of the normalized extrapolation distance. The confidence score remained close to unity within the training domain and decreased monotonically when the input conditions exceeded the data-supported range. This behavior reflects the dominant influence of the extrapolation distance on the overall confidence score, which is consistent with the weighting scheme adopted in this study.

Figure 7 shows a breakdown of the confidence components that contributed to the final confidence score. The distance-based component (S_dist) exhibited the largest variation across the extrapolation conditions, indicating that the extrapolation distance was the primary driver of confidence reduction. The statistical component (S_stat) provided a secondary contribution, reflecting the differences in fitting quality and predictive stability among the evaluated equations.

In contrast, the physics- and knowledge-graph-based components (S_phys) remained constant across all cases. Because all the equations were generated under MKG-based subgraph guidance and exhibited structural consistency with the Jacobs working curve formulation, no explicit violations of the encoded physical constraints were observed. Consequently, S_phys did not contribute to discrimination between the models in the present experimental setting.

Altogether, these results show that the final confidence score is governed primarily by the extrapolation distance and statistical stability, whereas physics-based consistency acts as a validity-preserving constraint rather than a differentiating performance metric in MKG-guided generation.

## 4. Discussion

### 4.1 Limitations and future work

This study has several limitations. First, the proposed framework was evaluated on a focused corpus of 20 papers and a single experimental dataset, focusing primarily on vat photopolymerization, and a representative physical model based on the Jacobian working curve. Although this choice allows for a controlled and interpretable evaluation of equation-centric knowledge extraction and extrapolation, it restricts the generalizability of the reported results to other additive manufacturing processes, materials, or multi-physics models.

Second, an extrapolative evaluation was conducted under a single out-of-range condition representing a 20% extension beyond the training domain. Although the proposed confidence scoring framework captures the relative changes in extrapolation risk, it does not constitute a comprehensive validation across a continuous extrapolation space or more extreme extrapolation distances. Therefore, the reported confidence scores should be interpreted as comparative indicators rather than as absolute measures of extrapolation validity.

Third, the physics and knowledge graph-based confidence components (S_phys) did not exhibit discriminative variation among the models evaluated in the present experiments. This behavior reflects the fact that all equations were generated under MKG-based subgraph guidance and therefore, satisfied the encoded physical constraints. Although this confirms that the knowledge graph effectively enforces physical consistency, it limits the physics-based score from differentiating between the models in the current experimental setting.

Fourth, the current framework focuses on single-equation extrapolation for a univariate input–output relationship. Extensions to multi-equation systems, coupled variables, and higher-dimensional parameter spaces remain unexplored.

Despite these limitations, this study serves as a proof-of-concept demonstrating how ontology-guided knowledge extraction, subgraph-conditioned equation generation, and confidence-aware extrapolation assessment can be integrated into a unified framework. Future work will extend this approach to broader process domains, more complex multi-variable equations, and richer sets of physics-informed constraints to further evaluate its scalability and generalizability.

# 5. Conclusions

This study proposes an ontology-guided, equation-centric knowledge extraction and extrapolative modeling framework that integrates LLM with an AM-MKG.

The results demonstrate that ontology-guided extraction significantly improves both the structural coherence and quantitative reliability of the extracted knowledge, thereby providing a robust foundation for downstream equation generation. When conditioned on MKG-based subgraphs, the LLM consistently generated equations that preserved physically meaningful functional structures and exhibited stable behavior in the extrapolation regimes, in contrast to the unstable and non-physical trends observed without structured guidance.

In addition to equation generation, this study introduces a confidence-aware extrapolation assessment framework that distinguishes between predictive uncertainty and extrapolation reliability. By integrating extrapolation distance, statistical stability, and knowledge graph-based physical consistency, the proposed confidence score provides a principled mechanism for comparing extrapolated predictions under limited data.

The proposed framework demonstrates how ontology-guided knowledge representation, subgraph-conditioned equation generation, and confidence-based extrapolation assessment can be coherently combined into a unified modeling pipeline. Although the current evaluation is limited to a focused dataset and a representative physical model, the results highlight the potential of knowledge graph augmented LLM as reliable tools for equation-centered reasoning and extrapolative modeling in additive manufacturing. Future studies will extend this framework to broader process domains, multi-variate systems, and richer physical constraints to further evaluate its generality and scalability.

## Conflicts of Interest

The authors declare no conflict of interest.

## Author Contributions

**Yeongbin Cha**: Conceptualization, Methodology, Data curation, Formal analysis, Visualization, Investigation, Writing – original draft. **Namjung Kim**: Conceptualization, Formal analysis, Writing – original draft, review and editing, Project administration, Funding acquisition.

## Funding

This work was supported by the National Research Foundation of Korea(NRF) grant funded by the Korea government (MSIT) (RS-2025-16069590).

## Data Availability

All data used and generated in this study are available in the article and Supplementary Information. Additional data related to this study are available from the corresponding author upon request.

# Acknowledgements

# Appendix A. Formal Competency Questions and Minimal AM-KG Ontology Specification

This appendix provides the formal competency questions (CQs) that guided the design of the minimal AM-KG ontology, together with the corresponding ontology schema specification. As described in Section 2.2, these competency questions serve as explicit functional requirements that determine the minimum set of entity types and relations required to support structured equation extraction, semantic validation, and physics-informed extrapolative modeling in the additive manufacturing domain.

The CQs are organized according to their intended inference roles, including equation structure interpretation, parameter–performance influence analysis, applicability and validity assessment, and semantic consistency evaluation. By explicitly linking each representational element of the ontology schema to one or more competency questions, this appendix provides a traceable and verifiable justification for the minimal ontology design adopted in this study.

## A.1 Competency questions for equation-centered knowledge representation

This subsection lists the competency questions grouped by their intended inference roles.

CQ-1. Equation structural queries

These questions ensure that equations are represented as structured and interpretable mathematical knowledge rather than as isolated text strings.

CQ-1.1: What are the input variables of a given equation?

CQ-1.2: What is the output variable of the equation?

CQ-1.3: What assumptions are required for the equation to hold?

These CQs require the ontology to explicitly represent equations, variables, and assumptions, together with their semantic relationships.

CQ-2. Parameter–performance influence queries

These questions enable the ontology to capture causal relationships between process parameters and performance metrics in additive manufacturing processes.

CQ-2.1: Which process parameters influence a given performance metric?

CQ-2.2: Is the direction of influence (positive or negative) consistent across literature sources?

These CQs support trend consistency analysis and constrain extrapolative modeling to physically plausible behaviors.

CQ-3. Applicability and validity queries (Extrapolation-Oriented)

These questions determine whether an equation can be validly applied under specific operating conditions.

CQ-3.1: What is the valid operating regime (parameter range) for a given equation?

CQ-3.2: Which equations in the knowledge graph can be used to predict a specific performance metric?

These CQs enable equation selection, dimensional consistency checking, and confidence-aware extrapolation.

CQ-4. Semantic consistency queries

These questions assess semantic alignment between mathematical expressions and their textual descriptions.

CQ-4.1: Do the mathematical expressions and textual descriptions yield consistent semantic conclusions?

CQ-4.2: What alternative equations describe the same physical phenomenon, and how do they differ structurally?

## A.2 Summary of ontology requirements

Together, the competency questions defined in this appendix specify the minimum functional capability required for (1) structured mathematical knowledge extraction, (2) physics-grounded equation selection, (3) knowledge-graph-enhanced extrapolation, and (4) consistency validation between equations, assumptions, and empirical descriptions.

## A.3 Minimal AM-KG ontology schema (Table A1)

**Table A.1. Minimal AM-KG Ontology Schema**

| Category | Name | Description |
|---|---|---|
| Entity | Equation | Mathematical expressions describing AM process behavior and performance relationships |
| | Variable | Input or output quantities appearing in equations, including process variables and material properties |
|  | Assumption | Physical, geometric, or modeling assumptions required for an equation to hold |
|  | Process Parameter | Controllable AM process parameters such as laser power, scan speed, or layer thickness |
|  | Performance | Target performance metrics such as strength, density, or surface quality |
|  | Regime | Valid operating ranges and boundary conditions |
|  | Material | Resin compositions and material components |
| Relation | has_input | Links an equation to its input variables |
|  | has_output | Links an equation to its output variable |
|  | influences | Represents causal influence of a process parameter on a performance metric, with directional sign |
|  | requires_assumption | Specifies assumptions required for an equation to be valid |
|  | valid_in_regime | Defines the operating or parameter regime in which an equation applies |
|  | corresponds_to | Mapping between process parameters and equation variables |
|  | uses_material | Materials use in a process |

## Appendix B. Ontology-Guided Preliminary Hint Structure

For each text chunk, the preprocessing stage generates ontology-aligned preliminary hint. This metadata provides a structured summary of potential entities and attributes contained within the chunk and is used as weak prior information to support subsequent LLM-based extraction. In particular, variables appearing in equations are tentatively categorized as inputs, outputs, constants, or unknowns based on their positions on the left-hand or right-hand side of the equation, while assumptions and regime-related information are organized into structured fields reflecting their relative criticality or numerical ranges.

These preliminary hints are organized into the following categories:

- Equation information: LaTeX-formatted equations accompanied by extracted variable lists. Variables are tentatively assigned roles (input, output, constant, or unknown) based on their positions within the equation.
- Process parameters: Candidate standardized manufacturing parameters such as exposure energy, laser power, and scan speed.
- Performance metrics: Candidate outcome variables including cure depth, tensile strength, porosity, and surface quality.
- Assumptions: Extracted assumption statements (e.g., steady-state conditions or negligible effects), each annotated with a coarse level of criticality.
- Regimes and ranges: Numerical ranges, limiting values, or recommended operating conditions identified from the local context.
- Materials: Mentions of materials appearing within the chunk.

Importantly, these preliminary hints do not directly instantiate nodes or relations in the knowledge graph. Instead, they serve as lightweight, ontology-aligned priors that guide the LLM during downstream extraction by supporting candidate validation and refinement.

# Appendix C. LLM Prompting Configuration for Knowledge Graph Construction

This appendix provides the prompt templates and output schemas used in the ontology-guided entity and relation extraction pipeline described in Section 2.4. Rather than repeating the methodological rationale presented in the main text, this appendix illustrates how the proposed approach is instantiated at the prompt level through representative structures and few-shot examples.

## C.1 Entity extraction prompt structure

- System Role Specification

The entity extraction prompt assigns the LLMs the role of an expert in additive manufacturing knowledge graph construction. The model is explicitly instructed to extract entities in strict accordance with ontology-defined types and naming conventions, and to include only entities that are supported by explicit textual evidence.

- Entity Type Definitions

Six entity types are defined for the entity extraction stage, each with specific formatting requirements:

- ProcessParameter: Controllable process parameters
- (e.g., laser_power, scan_speed)
- *Format*: lowercase, underscore-separated (snake_case)
- Performance: Measurable output performance metrics
- (e.g., tensile_strength, surface_roughness)
- *Format*: lowercase, underscore-separated (snake_case)
- Variable: Symbolic variables appearing in equations
- (e.g., P, v, E)
- *Format*: LaTeX-compatible notation
- Equation: Mathematical expressions representing physical relationships
- *Format*: descriptive identifier (e.g., equation_energy_density) accompanied by a complete LaTeX expression
- Assumption: Physical or experimental assumptions
- (e.g., steady_state, isothermal)
- *Format*: lowercase, underscore-separated
- Regime: Parameter range or operating condition constraints
- (e.g., laser_power_100_300_W)
- *Format*: parameter_min_max_unit
- Ontology-Guided Hint Injection

Ontology-guided preliminary hints generated during preprocessing are injected at the beginning of the prompt as a structured block. These hints include candidate process parameters, performance metrics, equations, assumptions, and regimes. They function as weak priors rather than enforced annotations. To prevent prompt overload, the total size of hints per category is capped at 2,000 tokens.

- Output Schema

Entity extraction outputs are returned in JSON format. Each extracted entity includes:

- entity name
- entity type
- short description
- unit (when applicable)

- evidence text chunk hash

## C.2 Few-Shot example for entity extraction

A representative few-shot example included in the entity extraction prompt is shown below.

1. Input Text (Excerpt)

The cure depth C_d increases logarithmically with exposure energy E,

following C_d = D_p ln(E / E_c),

where D_p and E_c are material-dependent constants.

2. Ontology-Guided Hints (Excerpt)

Process Parameters:

- exposure_energy

Performance Metrics:

- cure_depth

Equations:

- C_d = D_p ln(E / E_c)

Assumptions:

- steady_state

3. Example Output (JSON, Excerpt)

```
[
  {
    "name": "exposure_energy",
    "type": "ProcessParameter",
    "description": "Energy applied during exposure in photopolymerization",
    "evidence": "chunk_8f3a"
  },
  {
    "name": "cure_depth",
    "type": "Performance",
    "description": "Depth of cured material after exposure",
    "evidence": "chunk_8f3a"
```

```
    },
    {
      "name": "equation_cure_depth",
      "type": "Equation",
      "latex": "C_d = D_p \\ln(E / E_c)",
      "variables": [
        { "symbol": "C_d", "role": "output" },
        { "symbol": "E", "role": "input" },
        { "symbol": "D_p", "role": "constant" },
        { "symbol": "E_c", "role": "constant" }
      ],
      "evidence": "chunk_8f3a"
    },
    {
      "name": "steady_state",
      "type": "Assumption",
      "description": "The process is assumed to operate under steady-state conditions",
      "evidence": "chunk_8f3a"
    }
  ]
```

This example demonstrates how ontology-guided hints, source text, and schema constraints jointly guide the LLM toward evidence-grounded and ontology-compliant entity extraction.

## C.3 Connection to the main methodology

The prompt structure and few-shot example presented in this appendix operationalize the candidate proposal–verification strategy described in Section 2.4.2. By combining ontology-guided priors with explicit evidence-based constraints, the LLM functions as a controlled extractor rather than a free-form generator, ensuring consistency with the ontology and robustness in downstream knowledge graph construction.

# Appendix D. Source Papers Used for AM-MKG Construction

# Appendix E. Ontology-Constrained Equation Generation Algorithm

This algorithm summarizes the ontology-constrained procedure used to generate physically and semantically valid symbolic equations from a retrieved MKG-RAG subgraph.

---
**Algorithm E.1** Ontology-Constrained LLM-Based Equation Generation
---
**Require**: Retrieved MKG-RAG subgraph $G_{sub}$
  Input variable set $X = \{x_1, \cdots, x_m\}$
  Target variable $y$
  Number of candidates $M$

**Ensure**: Set of symbolic equation candidates $\{\hat{f}_i(X)\}_{i=1}^{M}$

1: Construct system prompt with physical and ontological constraints
2: Serialize $G_{sub}$ into structured textual context
3: Construct user prompt requesting $M$ equations of the form $y = f(X)$
4: Query LLM with JSON-only output constraint
5: Parse returned JSON and extract candidate equations
6: **for** each candidate equation **do**
7:   **if** the equation violates ontology constraints or output schema **then**
8:     Discard the candidate
9:   **else**
10:     Normalize the equation into symbolic form
---

11:   **end if**

12: **end for**

13: **return** ontology-grounded equation candidate